\documentclass[conference]{IEEEtran}

\usepackage[]{graphicx}
\usepackage{caption}
\usepackage{subcaption}
\usepackage{epstopdf}
\usepackage{cite}
\usepackage{textcomp}
\usepackage{cite}
\usepackage{graphicx}
\usepackage{amsmath}
\usepackage{amsfonts}
\usepackage{amsthm}

\usepackage{algorithm}
\usepackage[noend]{algpseudocode}
\DeclareMathOperator*{\argmin}{arg\,min} 
 
\DeclareCaptionFont{tiny}{\tiny}
\DeclareCaptionFont{small}{\small}
\DeclareCaptionFont{footnotesize}{\footnotesize}

\makeatletter
\def\BState{\State\hskip-\ALG@thistlm}
\makeatother

\ifCLASSINFOpdf
\else
\fi
\begin{document}
	%
	\title{Multiresolution Approach to Acceleration of Iterative Image Reconstruction for X-Ray Imaging for Security Applications}

	\author{\IEEEauthorblockN{Soysal Degirmenci}
		\IEEEauthorblockA{Department of Electrical and\\ Systems Engineering\\
			Washington University in St. Louis\\
			Saint Louis, Missouri 63130\\
			Email: s.degirmenci@wustl.edu}
		\and
		\IEEEauthorblockN{Joseph A. O'Sullivan}
		\IEEEauthorblockA{Department of Electrical and\\ Systems Engineering\\
			Washington University in St. Louis\\
			Saint Louis, Missouri 63130\\
			Email: jao@wustl.edu}
		\and
		\IEEEauthorblockN{David G. Politte}
		\IEEEauthorblockA{Mallinckrodt Institute\\ of Radiology\\
			Washington University School of Medicine\\
			Saint Louis, Missouri 63110\\
			Email: politted@wustl.edu}}
	
	
	%


	\maketitle
	
	\begin{abstract}
	Three-dimensional x-ray CT image reconstruction in baggage scanning in security applications is an important research field. The variety of materials to be reconstructed is broader than medical x-ray imaging. Presence of high attenuating materials such as metal may cause artifacts if analytical reconstruction methods are used. Statistical modeling and the resultant iterative algorithms are known to reduce these artifacts and present good quantitative accuracy in estimates of linear attenuation coefficients. However, iterative algorithms may require computations in order to achieve quantitatively accurate results. For the case of baggage scanning, in order to provide fast accurate inspection throughput, they must be accelerated drastically. There are many approaches proposed in the literature to increase speed of convergence. This paper presents a new method that estimates the wavelet coefficients of the images in the discrete wavelet transform domain instead of the image space itself. Initially, surrogate functions are created around approximation coefficients only. As the iterations proceed, the wavelet tree on which the updates are made is expanded based on a criterion and detail coefficients at each level are updated and the tree is expanded this way. For example, in the smooth regions of the image the detail coefficients are not updated while the coefficients that represent the high-frequency component around edges are being updated, thus saving time by focusing computations where they are needed. This approach is implemented on real data from a SureScan\textsuperscript{TM} \emph{x}1000 Explosive Detection System\footnote{SureScan\textsuperscript{TM} is a trademark of the SureScan Corporation.} and compared to straightforward implementation of the unregularized alternating minimization of O'Sullivan and Benac \cite{OSullivanBenac07}.
	\end{abstract}
	

	%
	\IEEEpeerreviewmaketitle

	\section{Summary}
	X-ray CT image reconstruction algorithms often are designed to trade-off a data fit term and an image roughness term. Typical examples of data fit terms are squared error\cite{thibault2007three} and log-likelihood.  Typical examples of image roughness terms are total variation, Gauss-Markov random field priors\cite{bouman1993generalized}, and Huber class roughness penalties.  A complementary approach for penalizing roughness is to represent the image using a wavelet (or other multiresolution) expansion, directly estimate the wavelet coefficients, and introduce a penalty (often $L_1$) on the wavelet coefficients\cite{ramani2012splitting,xu2014sparsity}. Multigrid approach\cite{MultigridSPIE04,pan1991numerical,oh2005general,oh2006multigrid} has been shown to be successful in some image reconstruction methods in terms of achieving faster convergence speed. The idea is to move through different grid levels over time. At any grid level, the voxel size is the same throughout the image domain. The approach we describe in this paper is closest to this with the following differences:
	\begin{itemize}
		\item Our data fit term is a Poisson log-likelihood with mean determined by Beer's law
		\item We use an alternating minimization framework to derive an algorithm that is guaranteed to decrease the cost function at every iteration
		\item We extend the prior alternating minimization framework to point spread functions with negative values
		\item We update only a subset of wavelet coefficients, constrained to be on a tree, thereby decreasing the computational complexity per iteration relative to fully updating the image
		\item We adaptively update the tree defining which wavelet coefficients are updated at each iteration
		\item Our wavelet tree structure results in image domain representation of voxels with different sizes
		\item We incorporate an adaptive threshold allowing the computation of a sequence of images of increasing resolution, with increasing roughness and increasing log-likelihood
	\end{itemize}
	The result is a fast, adaptive, iterative image reconstruction algorithm.
	\section{Introduction}
	The problem to be minimized for x-ray CT in this paper is penalized likelihood estimation with a Poisson log-likelihood data fitting term and a regularization term, optimized over image $\boldsymbol{\mu} \in \mathbb{R}^{N}_{+}$. It is shown in \cite{OSullivanBenac07} that maximization of the Poisson log-likelihood term is equivalent to minimizing the I-divergence\footnote{I-divergence between two vectors $ \boldsymbol{p}, \boldsymbol{q}\in\mathbb{R}^{N}_{+}$ is defined as $I(\boldsymbol{p}||\boldsymbol{q})=\sum_{i} p_{i}log(\frac{p_{i}}{q_{i}}) - p_{i} + q_{i}$.}  between the transmission data $\boldsymbol{d} $ and the estimated mean $\boldsymbol{q}(\boldsymbol{\mu}) \in \mathbb{R}_{+}^{M}$, where $\boldsymbol{q}(\boldsymbol{\mu})(y) = \boldsymbol{I_{0}}(y) \exp(-\sum_{x}h(y|x)\mu(x))$, $\boldsymbol{I_{0}}(.)$ is the incident photon count vector, $h(y|x)$ is an element of the system matrix $\boldsymbol{H} \in \mathbb{R}^{M\times N}_{+} $ that represents the length of the intersection between the ray path of index $y \in \mathcal{Y}^{M}$ and voxel of index $x \in \mathcal{X}^{N}$. 
	Then this penalized likelihood estimation problem can be formulated as \cite{SPIE2015}

	\begin{equation} \label{eq:1.1}
	\boldsymbol{\mu}^{*}_{PML} = \argmin_{\boldsymbol{\mu} \ge 0} I(\boldsymbol{d}||\boldsymbol{q}(\boldsymbol{\mu}) ) + \lambda R(\boldsymbol{\mu}),
	\end{equation}
	where $R(\boldsymbol{\mu})$ is a regularization term selected as a roughness penalty and $\lambda \ge 0$ is the parameter that controls the level of roughness imposed on the image. Also, it is important to note that the non-negativity constraint on $\boldsymbol{\mu}$ is due to the nature of linear attenuation coefficients of materials. Since there is no closed form solution to this problem, we solve it iteratively. At each iteration, a surrogate function that approximates the original objective function is minimized, which in turn decreases the original objective function. In our recent work   \cite{SPIE2015}, we generalized the formulation of surrogate functions in \cite{OSullivanBenac07} for data fitting term to the regularization term. The idea is to use Jensen's inequality to decouple the objective function and form many one-parameter convex functions, minimize them, and iterate.
	
	Assume that there exists a discrete wavelet inverse transform matrix $\boldsymbol{\Omega} \in \mathbb{R}^{N \times N}$ that is non-singular. Then, the image $\boldsymbol{\mu}$ can be represented as
	\begin{equation} \label{eq:1.2}
	\boldsymbol{\mu}=\boldsymbol{\Omega}\boldsymbol{\beta},
	\end{equation}
	where $\boldsymbol{\beta}$ is the vector of wavelet coefficients. The problem in this paper can then be written as
	\begin{eqnarray} \label{eq:1.3}
	\boldsymbol{\beta}^{*}_{PML} = \argmin_{\boldsymbol{\beta}} I(\boldsymbol{d}||\boldsymbol{q}(\boldsymbol{\Omega}\boldsymbol{\beta}) ) + \lambda R(\boldsymbol{\Omega}\boldsymbol{\beta}) \\ 
	\text{ subject to } \boldsymbol{\Omega}\boldsymbol{\beta} \ge 0 \nonumber
	\end{eqnarray}
	
	Below, the derivation of the surrogate functions for the data fitting term is shown. A similar approach yields surrogate functions for the regularization term as well.

	The I-divergence term can be written as
	\begin{align} \label{eq:1.4}
	I(\boldsymbol{d}||\boldsymbol{q}(\boldsymbol{\Omega}\boldsymbol{\beta}))  &= \sum_{y} d(y) \sum_{x} h(y|x) \sum_{z} \omega(z|y) \beta(z) \nonumber \\
	&+ \sum_{y} I_{0}(y) \exp \big(-\sum_{x} h(y|x) \sum_{z} \omega(x|z) \beta(z) \big) \nonumber \\
	&+ constant(y).
	\end{align}
	
	For simplicity, define the matrix $\boldsymbol{\Phi} =  \boldsymbol{H}\boldsymbol{\Omega}$, where $\phi(y|z)$ is the system matrix element between ray path of index $y$ and wavelet coefficient of index $z \in \mathcal{Z}^{N}$. Assume that there exists a known estimate $\boldsymbol{\hat{\beta}}$ and $\boldsymbol{\hat{q}}(y) = \boldsymbol{I_{0}}(y) \exp (-\sum_{x} h(y|x) \sum_{z} \omega(x|z) \hat{\beta}(z)) = \boldsymbol{I_{0}}(y) \exp (-\sum_{z} \phi(y|z) \hat{\beta}(z)) $. The terms in the I-divergence that depend on $\boldsymbol{\beta}$ are used to construct surrogate functions as follows.
	\begin{align} \label{eq:1.5}
	&= \sum_{y} d(y) \sum_{z} \phi(y|z) \beta{(z)}\nonumber \\
	&\quad \quad +\sum_{y} \hat{q}(y) \exp \big(-\sum_{z} \phi(y|z) (\beta(z) - \hat{\beta}(z) \big) \nonumber \\
	&\le \sum_{z} b(z) \beta{(z)} \nonumber\\
	&\quad \quad +\sum_{y} \sum_{z} \hat{q}(y) r(z|y) \exp(-\frac{\phi(y|z)}{r(z|y)} (\beta(z) - \hat{\beta}(z))),
	\end{align}
	where
	\begin{equation} \label{eq:1.6}
	b(z) = \sum_{y} d(y)\phi(y|z),
	\end{equation}
	the convex decomposition lemma~\cite{OSullivanBenac07} is used for $r(z|y) \ge 0$, $\sum_{z} r(z|y) \le 1$. $r(z|y)$ can be chosen as
	\[
	r(z|y) =
	\begin{cases}
	\frac{|\phi(y|z)|}{Z_{0}},& \text{ if }  z \in \mathcal{Z}_{s} \\
	0,& \text{ if } z \notin \mathcal{Z}_{s},
	\end{cases}
	\]
	\begin{equation} \label{eq:1.61}
	Z_{0} = \max_{y} \sum_{z \in \mathcal{Z}_{s}} |\phi(y|z)|,
	\end{equation}
	and $\mathcal{Z}_{s} \subseteq \mathcal{Z}$, $\mathcal{Z}_{s} \neq \emptyset$.\footnote{$\mathcal{Z}_{s}$ represents a subset of the wavelet domain to be chosen for update. In our approach, we choose it in a way that every voxel in image domain is represented at any iteration, possibly with different numbers of coefficients. This subset can be fixed or be varied over iterations.}
	\begin{align} \label{eq:1.7}
	&\le \sum_{z \in \mathbb{Z}_{s}} b(z) \beta{(z)} \nonumber \\
	&\quad \quad +\sum_{y} \sum_{z \in \mathcal{Z}_{s}} \hat{q}(y) \frac{|\phi(y|z)|}{Z_{0}} \exp(-Z_{0}\frac{\phi(y|z)}{|\phi(y|z)|} (\beta(z) - \hat{\beta}(z))) \nonumber \\
	&\quad \quad +\sum_{z' \notin \mathcal{Z}_{s}} const(z')
	\end{align}
	Adding the constant term in I-divergence, we define our surrogate function,
	\begin{multline} \label{eq:1.8}
	\hat{I}_{\mathcal{Z}_{s}}(\boldsymbol{d}||\boldsymbol{q};\boldsymbol{\beta},\boldsymbol{\hat{\beta}}) = \sum_{z \in \mathcal{Z}_{s}} b(z) \beta{(z)}\\
	+\sum_{y} \sum_{z \in \mathcal{Z}_{s}} \hat{q}(y) \frac{|\phi(y|z)|}{Z_{0}} \exp(-Z_{0}\frac{\phi(y|z)}{|\phi(y|z)|} (\beta(z) - \hat{\beta}(z))) \\
	+\sum_{z' \notin \mathcal{Z}_{s}} const(z') + const(y)
	\end{multline}
	It is clear to see that this is a one-parameter convex function over each $\beta(z)$ and the gradient with respect to $\beta(z)$ is given as:
	\begin{eqnarray} \label{eq:1.9}
	\hat{I}'_{\mathcal{Z}_{s}}(\boldsymbol{d}||\boldsymbol{q};\boldsymbol{\beta},\boldsymbol{\hat{\beta}})&=& b(z) - \hat{b}_{+}(z)\exp(-Z_{0}(\beta(z)-\hat{\beta}(z))) \nonumber \\ 
	&-& \hat{b}_{-}(z)\exp(Z_{0}(\beta(z)-\hat{\beta}(z)))
	\end{eqnarray}
	where
	\begin{align} \label{eq:1.10}
	\hat{b}_{+}(z) &= \sum_{y, \phi(y|z)>0} \hat{q}(y)\phi(y|z),\\
	\hat{b}_{-}(z) &= \sum_{y, \phi(y|z)<0} \hat{q}(y)\phi(y|z).
	\end{align}
	The first-order necessary condition for a minimizer is to find the $\beta(z)$ for which the gradient is zero, which has a closed form solution. The algorithm is shown below. 
	
	\begin{algorithm}
		\caption{Unregularized Wavelet AM Algorithm}\label{unr-wam}
		\begin{algorithmic}
			\State{\textbf{Inputs}: $\boldsymbol{\beta^{(0)}}, \boldsymbol{d}, \boldsymbol{I_{0}}, \boldsymbol{H}, \boldsymbol{\Phi}, \boldsymbol{\Omega}, \mathcal{Z}^{(j)}_{s} \text{for} j=0, 1, ..., (J-1) $}
			\State {Precompute $b(z) =  \sum_{y} d(y) \phi(y|z) $}

			\For{$j=0, 1, ..., (J-1)$}
			\State $\hat{q}^{(j)}(y) = I_{0}(y) \exp(-\sum_{z \in \mathcal{Z}^{(j)}_{s}} \phi(y|z) \hat{\beta}^{(j)}(z))$
			\State {$ Z_{0}^{(j)} = \max_{y} \sum_{z \in \mathcal{Z}^{(j)}_{s}} |\phi(y|z)|$}
			\For {every $z \in \mathcal{Z}^{(j)}_{s}$}
			\State $\hat{b}^{(j)}_{+}(z) = \sum_{y, \phi(y|z)>0} \hat{q}(y)\phi(y|z)$
			\State $\hat{b}^{(j)}_{-}(z) = \sum_{y, \phi(y|z)<0} \hat{q}(y)\phi(y|z)$
			\State $\hat{\beta}^{(j+1)}(z) = \tilde{\beta}(z)$ where  
			\State $b(z) - \hat{b}^{(j)}_{+}(z)\exp(-Z_{0}^{(j)}(\tilde{\beta}(z)-\hat{\beta}^{(j)}(z)))$
			
		\State{$- \hat{b}^{(j)}_{-}(z)\exp(Z_{0}^{(j)}(\tilde{\beta}(z)-\hat{\beta}^{(j)}(z))) = 0$}
			\EndFor
			\State{\textbf{end for}}
			\EndFor
			\State{\textbf{end for}}
						
		\end{algorithmic}
	\end{algorithm}	
	
	\section{Results}
\begin{figure}[!t]
	\centering
	\includegraphics[width=0.9\linewidth]{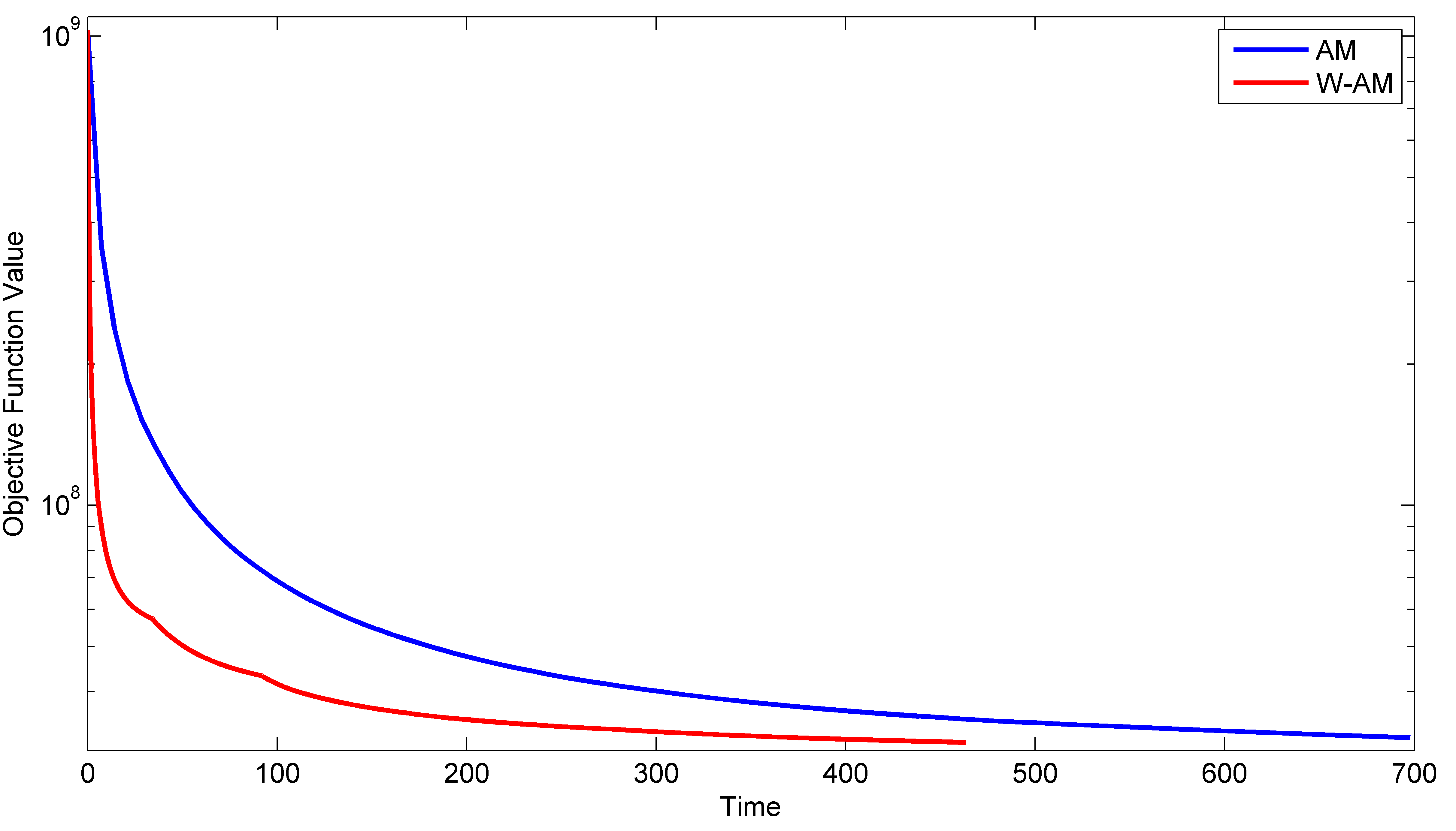}
	\caption{Objective function values vs. time for AM and Wavelet AM.}
	\label{fig_1}
	\end{figure}

\begin{figure}[!t]
		\centering
		\includegraphics[width=1.0\linewidth]{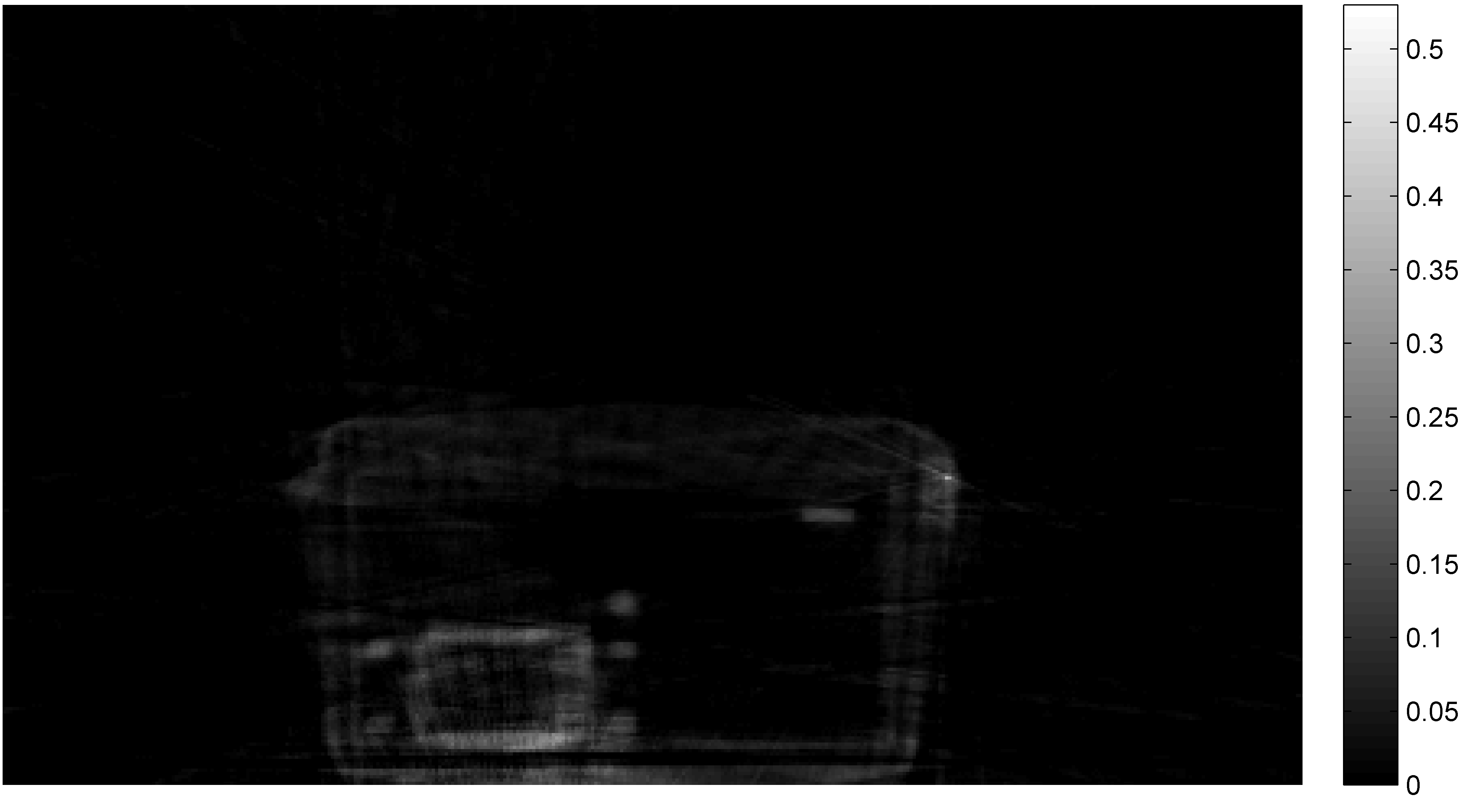}
		\caption{Image reconstructed with unregularized AM after 100 iterations.}
		\label{fig_2}
	\end{figure}

\begin{figure}[!t]
		\centering
		\includegraphics[width=1.0\linewidth]{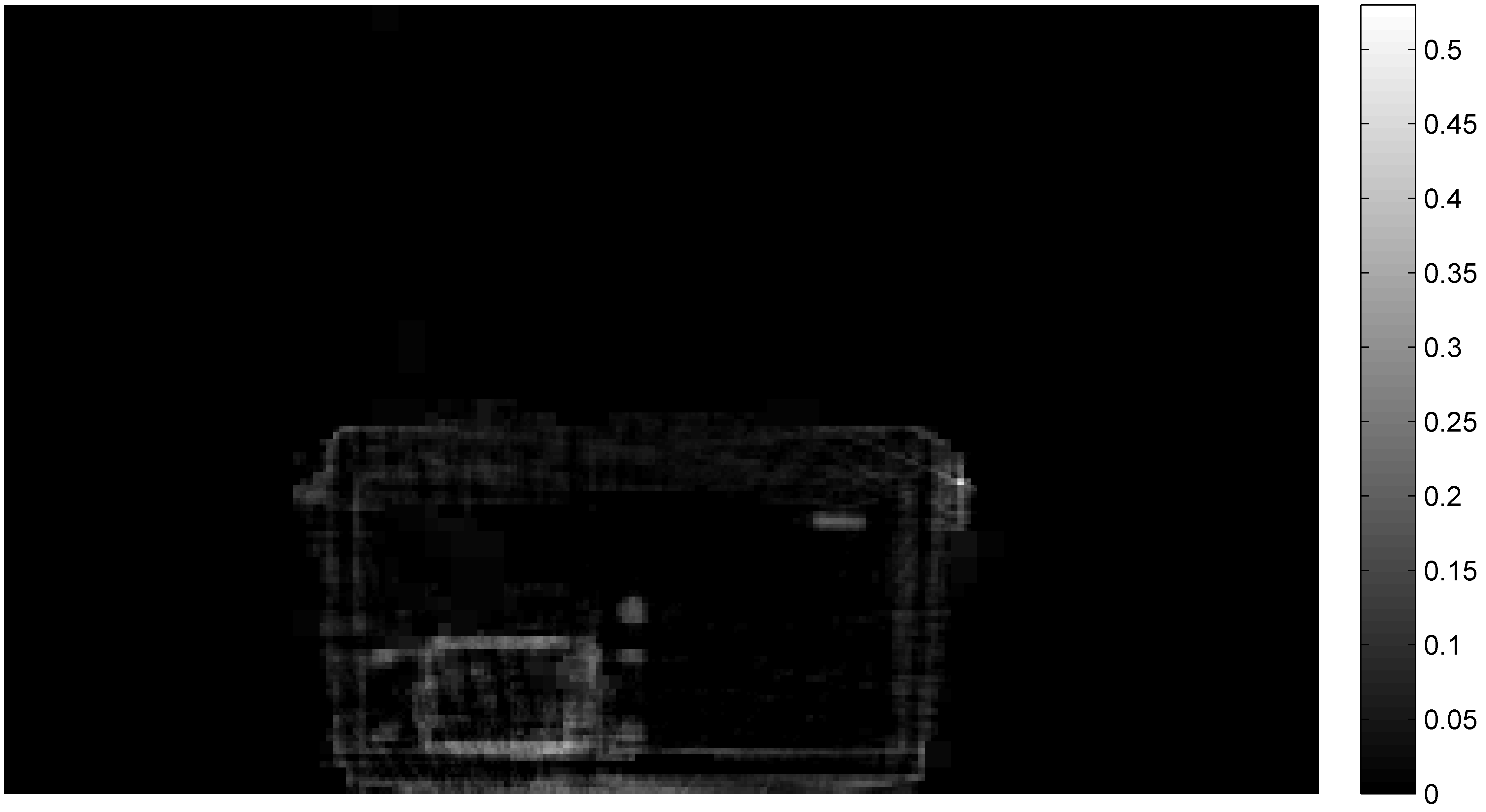}
		\caption{Image reconstructed with wavelet AM after 300 iterations.}
		\label{fig_3}
	\end{figure}

\begin{figure}[!t]
		\centering
		\includegraphics[width=1.0\linewidth]{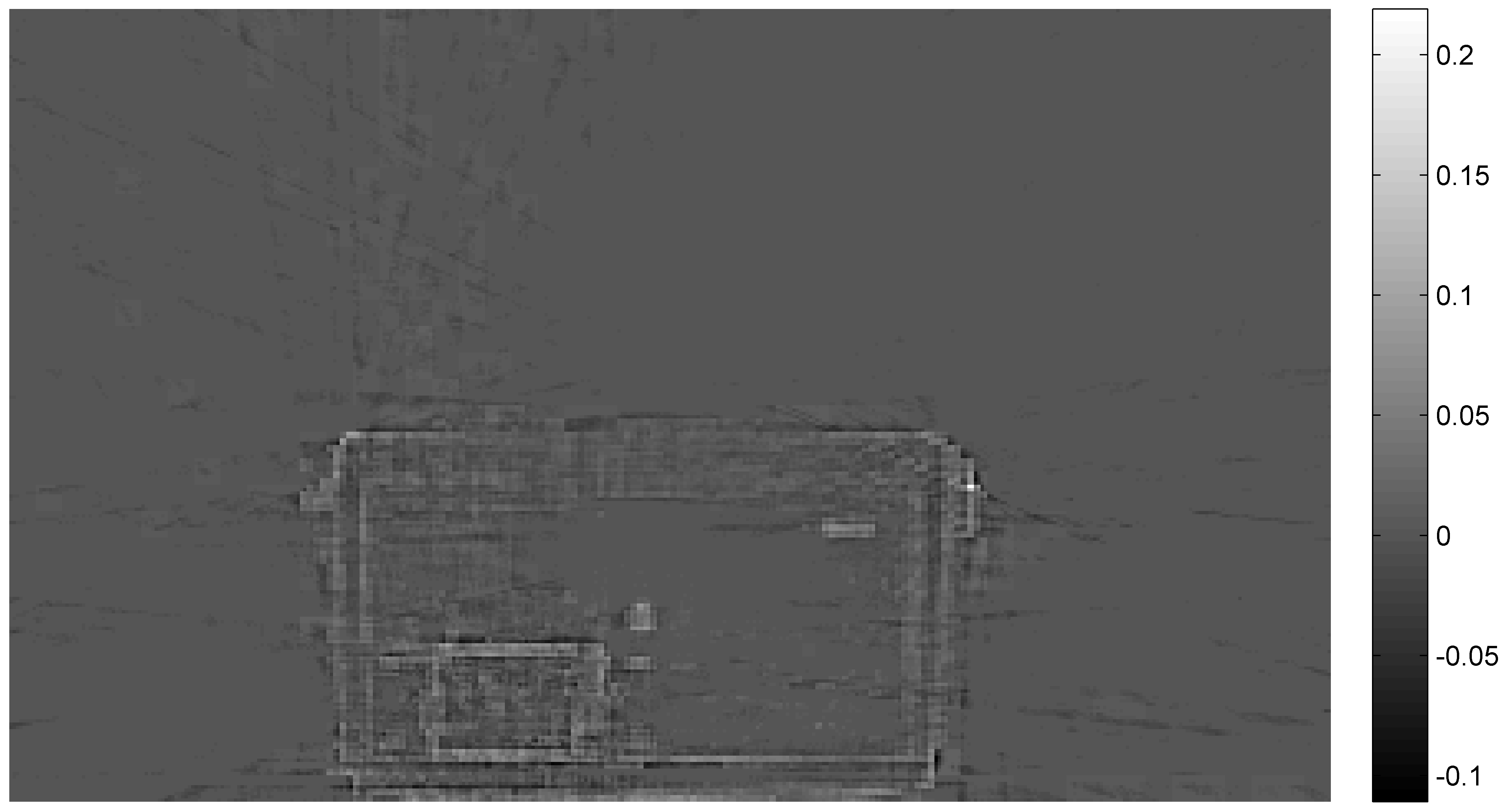}
		\caption{Difference image, unregularized AM image subtracted from wavelet AM image.}
		\label{fig_4}
	\end{figure}

	The multiresolution technique has been evaluated using a real data scan of the NIST Phantom Test Article A~\cite{NIST} acquired on a SureScan\textsuperscript{TM} \emph{x}1000 Explosive Detection System. A two dimensional Level 3 Haar disrete wavelet transform is used to represent each z-slice of the three dimensional image domain. The wavelet tree, $\mathcal{Z}^{(j)}_{s}$ is initialized to consist of approximation coefficients only. At iteration number $64$, the coefficients are back projected to the image space, voxel values across z-slices are summed up and the pixels whose values were larger than $0.1$ times the maximum of the summed image were chosen to expand one level. Then, at iteration number $128$, the same procedure is applied with the same factor to expand one level further, and the last expansion is done at iteration number $256$. Figure \ref{fig_1} shows objective function values versus time for unregularized alternating minimization algorithm (AM)\cite{OSullivanBenac07} and unregularized wavelet AM represented in this paper. AM algorithm has been run for 100 iterations while Wavelet AM has been run for 300 iterations. Figures \ref{fig_2} and \ref{fig_3} show image slices reconstructed from two algorithms at the same objective function value level. The difference between these two images (unregularized AM image subtracted from wavelet AM image) is shown in Figure \ref{fig_4}. It is important to note that even though two images are at the same objective function value level, the image reconstructed using wavelet AM has sharper edges. 
		

	\section{Conclusion}
	A fast, iterative, and adaptive algorithm for x-ray imaging was formulated and presented by using alternating minimization framework. The algorithm is guaranteed to decrease at each iteration and adaptive wavelet tree structure provides better utilization of computations. In other words, more computations are used for the regions with high frequency components like edges while less are used for smoother areas. The wavelet tree expansion used to reconstruct the image shown in the results section is one of many possible methods to perform it. Different ways to expand the tree will be investigated in the future. Different scale levels of discrete wavelet transform, different wavelet types and exploration of regularization are other parts to be explored later. Furthermore, this method can be combined with other acceleration methods like ordered subsets \cite{ErdoganFessler99}. Preliminary studies combining ordered subsets and wavelet AM showed promising results and will be investigated further.


	\section*{Acknowledgment}
	We thank Carl Bosch, Nawfel Tricha, Sean Corrigan, Michael Hollenbeck, and Assaf Mesika of SureScan\textsuperscript{TM} Corporation for their contributions to the collection, formatting, and sharing of the data.

	
	
	%
	\bibliography{bibtex_fully3d}   
	\bibliographystyle{IEEEtran}
	%
	%

\end{document}